\documentclass{mva_style}
\usepackage{graphicx}
\usepackage{caption}
\usepackage{float}
\usepackage{subcaption}
\finalcopy 	

\begin{document}
\title{Improving image classifiers for small datasets \\ by learning rate adaptations}

\author{
  Sourav Mishra\\
 The University of Tokyo\\
  7-3-1 Hongo, Bunkyo-ku, Tokyo\\
  {\tt sourav@ay-lab.org}\\
  \and
 Toshihiko Yamasaki\\
  The University of Tokyo\\
  7-3-1 Hongo, Bunkyo-ku, Tokyo\\
  {\tt yamasaki@ay-lab.org}\\
  \and
  Hideaki Imaizumi\\
  exMedio Inc.,\\
  3-5-1 Kojimachi, Chiyoda-ku, Tokyo\\
 {\tt imaq@exmed.io}\\
}
\maketitle

\section*{\centering Abstract}
\textit{Our paper introduces an efficient combination of established techniques to 
    improve classifier performance, in terms of accuracy and training time. We achieve 
    two-fold to ten-fold speedup nearing state of the art accuracy, over different 
    model architectures by dynamically tuning the learning rate. We find it especially 
    beneficial in the case of a small dataset, where reliability of machine reasoning 
    is lower. We validate our approach by comparing our method versus vanilla training 
    on CIFAR-10. We also demonstrate its practical viability by implementing on an 
    unbalanced corpus of diagnostic images.}

\section{Introduction}
Image classification via deep learning has seen rapid improvements in the last few 
years. First introduced as a visual recognition challenge, its scope has permeated 
from industrial applications to medical uses \cite{russakovsky2015}. Various new 
architectures have been proposed since the introduction of AlexNet in 2012 
\cite{krizhevsky2012}, such as ResNet, DenseNet  and Inception \cite{kaiming2016, 
huang2017,szegedy2015}. These architectures have become the mainstay in most 
computer vision applications, as per current literature. 

In addition to emergence of new architectures, we have also observed a steady 
increase in the accuracy of classification reported  on standard datasets. However, 
this improvement cannot be solely attributed to the architectures alone. Learning 
paradigms such as newer loss functions, optimization methods such as 
Dropout~\cite{srivastava2014}, and DropBlock~\cite{ghiasi2018} along with better 
pre-processing routines have contributed to higher accuracy. However, some bottlenecks 
still remain. Conventional methods of training models require a large amount of data, 
often augmented or imputed from the original dataset to make it balanced. The phase 
of training also happens to be the longest task in the deep learning process. With 
the advent of cloud computing, much of the machine learning tasks have shifted to 
portals such as Amazon Web Services and Microsoft Azure. Larger architectures usually 
imply longer running tasks translating to higher billed costs. This is of paramount 
importance to research groups or startups working on limited time and financial 
resources to deliver outcomes.

To alleviate this training performance bottleneck, a large number of workarounds 
have been proposed over the years, albeit with lesser attention. In this paper, we 
have proposed an organic combination of existing techniques and some derivative ideas, 
focused on the hyperparameter of learning rate. We allow it to adapt predictably 
over the training phase to improve time (measured as wall time) as well as accuracy 
(measured as validation accuracy). Our empirical evaluations have led us to conclude 
that learning rate adaptation is an under-exploited resource in making a difference 
on the overall model performance.

In this paper we have presented the following:

\begin{itemize}
	\item We investigated optimizing small dataset training by combining cosine rate annealing, 
	cycle length multiplication and differential learning rates. 
	\item We benchmark our scheme against conventional training paradigms, using CIFAR-10 dataset
	and quantify the speedup factor.
	\item We demonstrate the viability of this method on a corpus of dermatological diagnostic images 
	which has an unbalanced nature.
\end{itemize}

Our paper is organized as follows. Following this \textit{Introduction} we elaborate on the 
\textit{Methods} in Section~2, describing the \textit{modus operandi} of baseline determination, 
modified scheme and benchmarking. We present the results in Section~3 and demonstrate a practical
application in Section~4. We conclude with Section~5 containing discussion \& inferences drawn.

\section{Methods}
\subsection{Baseline Preparation}
For computing the baseline, we chose multi-class classification on the CIFAR-10 dataset 
\cite{cifar10}. Our rationale of this choice was to select a data corpus reported commonly 
in contemporary literature, exhibiting sufficient variety and yet smaller than ImageNet 
or CIFAR-100 \cite{russakovsky2015}. Our classifier was built on PyTorch v0.4 framework 
with commonly recommended practices such as dynamic augmentation, early stopping and an 
option to resume training with a different (but fixed) learning rate. Pretrained ResNet-34, 
ResNet-50, ResNet-101, ResNet-152 and DenseNet-161 were chosen as the candidate architectures; 
a single GPU (NVIDIA Titan Xp 12 GB HBM2) was utilized for all the measurements. 
The batch size was set at 32.

Prior to model learning, we normalized the data with the recommended mean 
(0.4914, 0.4822, 0.4465) and standard deviation (0.2023, 0.1994, 0.2010). The images 
were split in ratio of 5:1 into training and validation set. We performed dynamic 
in-memory augmentation by crop, horizontal \& vertical flips. Additionally we 
introduced other augmentation methods such as random-zoom by the \textit{imgaug} 
package from Python repository. 

All the models were trained to at least 90\% validation accuracy for fair comparison. 
The training was commended with learning rate $\alpha=0.01$ and restarted manually 
with lower value ($\alpha=0.001$), whenever early stopping forced the training to halt. 
Setting sufficient number of epochs to train was chosen based on prior experience. 
The accuracy of models and time to train are elucidated in Table 1. 

\subsection{Finding Initial Learning Rate}
To optimize training performance, we focused on keeping learning rate $\alpha$ suitable 
throughout the model learning. Conventional wisdom dictates that learning rates should 
monotonically decrease during the course of training. However, starting with a value significantly 
smaller than theoretical optimum could lead to cost function never converging towards global minimum; 
similarly a larger value will lead to divergence or over-fitting. It led to us to the pertinent question 
of determining an optimal initial learning rate. To accomplish this objective, we introduced an initial 
rate finder described by Smith et al. \cite{smith2017}. The implementation used several mini-batches 
with gradually increasing values of $\alpha$, until the loss computed at end of each batch started 
decreasing dramatically. Finding the rate of change of loss, we could zero down to a good learning rate to 
begin. Figures 1 and 2 illustrate the learning rate range test for DenseNet-161.

\begin{figure}[t]
	\includegraphics[width=\linewidth]{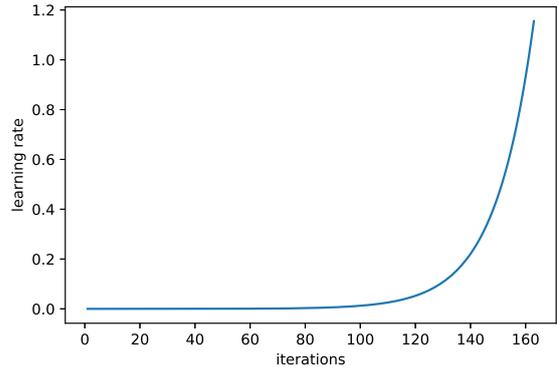}
	\caption{Determination of most suitable initial learning rate. The learning rate is systematically 
		increased over a large range to determine where the losses start reducing.}
	\centering
\end{figure}

\begin{figure}[t]
	\includegraphics[width=\linewidth]{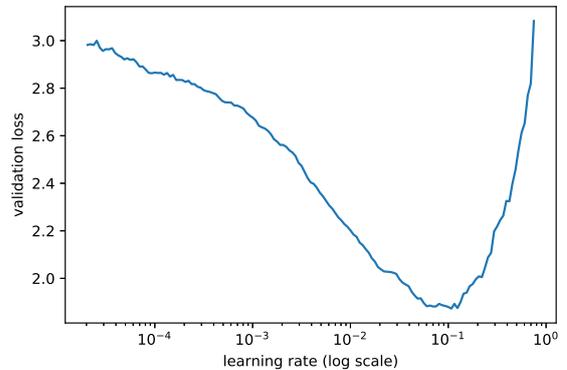}
	\caption{Plot of the losses measured over test mini-batches to determine the optimum initial 
		learning rate.}
	\centering
\end{figure}

\subsection{Cosine Rate Annealing}
Following determination of the best initial rate, we chose to train the network with 
transforms similar to our baseline measurement. The differences we introduced at this 
stage were to employ L2 regularization and running a pass through our data to pre-compute 
activation values for the final layer, keeping all other layers frozen. As the first step, we
froze all layers except the final layer to retain the complex features from ImageNet training. 
Initial shaping of the final layer promised performance advantages. Rather than keeping the 
learning rate fixed from what was determined, we adopted Stochastic Gradient Descent 
with Restarts (SGD-R), where cosine rate annealing gradually decreased the learning rate 
over the epoch from a designated value to zero \cite{Loshchilov2017}. The scheduling is 
governed as shown in Equation~1:

\begin{equation}
	\nu_t = \frac{1}{2} \left( 1 + \nu cos\left( \frac{t\pi}{T}\right) \right) + \nu_{min}
\end{equation}

\noindent where $\nu$ is the initial learning rate, $t$ is the iteration over the epoch, 
and $T$ is the total number of iterations to cover a epoch. Up to 10 epochs were run using 
this method, until the validation accuracy stabilized. This scheduling operation is 
illustrated by Figure 3 for a training task carried out on DenseNet-161.

\begin{figure}[t]
	\includegraphics[width=\linewidth]{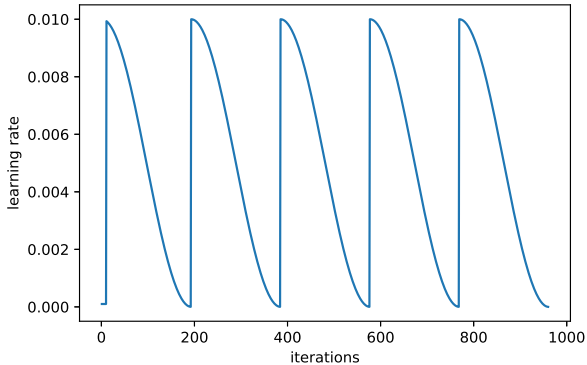}
	\caption{Plot of the cosine rate annealing over each epoch. The learning rate decreases from the 
	precomputed optimal value to a minimum, over the course of epoch only to again restart with a new epoch.}
	\centering
\end{figure}

\subsection{Cycle Length Multiplication and Differential Learning Rates}
Our final step in the optimization scheme was to unfreeze all the network layers, and use 
differential learning rates for separate sections of the networks. Before diving into 
the rationale, we explain the following two concepts with brevity:

\begin{description}
\item[Differential Learning Rate (DLR)]
  involved the process of assigning three separate learning rates, spanning the length of network \& 
  adhering to cosine rate annealing over the cycle length.   
 \item[Cycle Length Multiplication (CLM)] involved extending the cosine annealing over progressively 
 more integral number of  epochs (with each subsequent cycle), as described in \cite{Loshchilov2017}. 
\end{description}

Initial layers of convolutional neural networks (CNN) typically capture rudimentary features, with complexity 
of  the same increasing in later layers. More volatility can be envisioned  in higher layer with each pass 
of stochastic gradient descent (SGD). Therefore assigning a very low-rate of $\alpha=0.0001$, left the 
initial layers virtually undisturbed, whereas a moderately high rate of $\alpha=0.01$ allowed elasticity for 
change in the higher layers. The mid-section of the CNN was assigned a rate of $\alpha=0.001$. 

Concurrent with changing the designated learning rates over different sections, we also extended the 
cycle length with a multiplication factor of 2. As the model trained 
closer towards global minima, the parameters needed less perturbations. Gradually slowing down the 
SGD-R process guaranteed a better convergence towards global minima. The aforementioned step is 
illustrated by Figure 4, implemented on DenseNet-161.

\begin{figure}[t]
	\includegraphics[width=\linewidth]{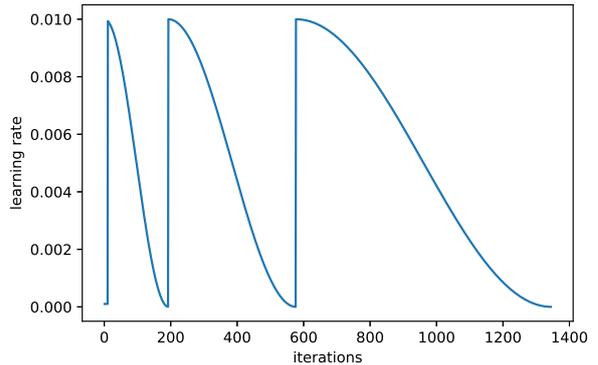}
	\caption{Cosine rate annealing extends to progressively higher number of 
	epochs in cycle length multiplication.}
	\centering
\end{figure}

\section{Results}
The results from conventional training, described in Section~2.1, are illustrated in Table 1. 
For all the candidate architectures, early stopping halted model learning before empirically 
assigned maximum epochs. As stated previously, the training was resumed with $\alpha=0.001$, 
with a target to achieve a minimum of 90\% validation accuracy. We have recorded learning duration 
and stable validation accuracy attained by early stopping with both the learning rates. Total time, 
which is a sum total of all learning durations, is presented in the last column.

\begin{table*}[!ht]
  \caption{Training on CIFAR-10 with conventional method}
  \vspace{-0.6em}
  \begin{center}
    \begin{tabular}{c | c c | c c | c}
      \hline
      \hline
      \makebox{Model} & {\small Accuracy ($\alpha$=0.01)} & {\small Time (s)} & {\small Accuracy ($\alpha=0.001$)} & {\small Time (s)}& {\small Total time (s)} \\
      \hline
      ResNet-34		&	86.25\% 	&	14749   &	90.36\%     &	3008    &   17757\\
      ResNet-50		&	86.56\% 	&	32596   &   90.54\%	    &	1442    &   34039\\
      ResNet-101	&	86.35\% 	&	58315   &	90.71\%     &	2323    &   60639\\
      ResNet-152	&	86.42\% 	&	88520   &   90.68\%	    &   3367    &	91888\\
      DenseNet-161	&	89.88\% 	&	51109   &	93.02\%	    &   3518    &	54628\\
      \hline
      \hline
    \end{tabular}
    \label{sample-table}
  \end{center}
\end{table*}

The results from our optimization scheme are illustrated in Table 2. The columns highlight wall time recorded for
phases described in Sections~2.2 through 2.4. The speedup factor on total time is indicated in the last column.

\begin{table*}[!ht]
	\caption{Training on CIFAR-10 with optimization scheme}
	\vspace{-1.5em}
	\begin{center}
		\begin{tabular}{c | c c | c c| c | c}
			\hline
			\hline
			\makebox{Model} & {\small Acc. with SGD-R} & {\small Time (s)} & {\small Acc. with DLR+CLM} 
			& {\small Time (s)} & {\small Total time} & {\small Speedup factor}\\
			\hline
			ResNet-34		&	82.45\%	&	3816	& 96.84\%	& 5840 & 9565  & 1.84   \\
			ResNet-50		&	80.34\%	&	5345	& 96.82\%	& 6472 & 11817 & 2.88   \\
			ResNet-101		&	84.20\%	&	2676	& 97.61\%	& 3998 & 6673  & 9.09   \\
			ResNet-152		&	82.50\%	&	3517	& 97.78\%	& 5496 & 9012  & 10.20  \\
			DenseNet-161	&   82.89\%	&   2001	& 97.15\%	& 5195 & 7195  & 7.59   \\
			\hline
			\hline
		\end{tabular}
		\label{cifar10-optim}
	\end{center}
\end{table*}

A confusion matrix for evaluation of the validation process on DenseNet-161 is presented in Figure 5. 
Matrices for other architectures are in the Appendix. We offer a few observations at this stage. 
We can see that models tune to near state-of-the-art accuracy values, without a high training 
time. Furthermore, they are capable of learning without over-fitting, on an orthodox 
choice of 5:1 training validation split of CIFAR-10. This demonstrates that learning rate is 
effective tool even with a smaller corpus of images. Most importantly, we observe that larger 
architecture receive better performance gains than smaller ones. We believe this is a valuable 
trait of our scheme, since current research is progressively moving towards using larger quantities 
of data and bigger architectures. 

\begin{figure}[t]
	\includegraphics[width=\linewidth]{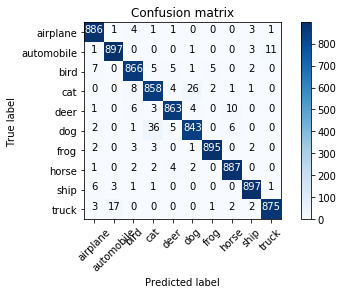}
	\caption{Confusion matrix for the quality of evaluation for CIFAR-10 on DenseNet-161}
	\centering
\end{figure} 

\section{Application}
To investigate benefits of our optimization technique and draw inferences, we introduce an application 
based on medical diagnostics for which this scheme was envisioned. 

Dermatological diseases exhibit a wide variety in their manifestation. At a time when demand for 
medical consultation is rising, there is a severe under-supply of dermatologists in many countries. 
The number of practitioners in US has plateaued at 3.6 per 100,000 people and several East Asian 
countries advocate mobile clinics or tele-medicine \cite{kimball2008}. In the absence of immediate 
avenues, people resort to general practitioners (GP). Statistics indicate that the opinions of GPs are 
concurrent with dermatologists only 57\% of the time \cite{lowell2001}. We attempt to provide machine 
learning based solutions to bridge this gap with a two-fold aim: To reduce workload of the dermatologist 
by aiding faster screening \& provide customized means to people for detecting possible skin problems

Most open medical image databases are small and unbalanced. Hence, there is a need to develop 
paradigms for effective training with limited information. Further, since disease labels could 
get updated periodically, or model requires customization for a new population type, training 
and deployment needs to be fairly rapid. Our schema manages to fulfill both the objectives, 
guaranteeing rapid deployment to the production servers. We chose a dermatological dataset 
with 10 unbalanced classes, containing 7543 images. The train-validation split was done unevenly 
to mimic a real database. We trained aforementioned architectures with vanilla scheme, followed 
by our optimization trick. We have compared the stable validation accuracy and total time across 
different architectures in Table 3. We again observed improved accuracy by around 10\% points 
and speedup between 3.1 and 5.7 times. A confusion matrix based on ResNet-152 is 
shown in Figure~6.

\begin{figure}[!htbp]
	\includegraphics[width=\linewidth]{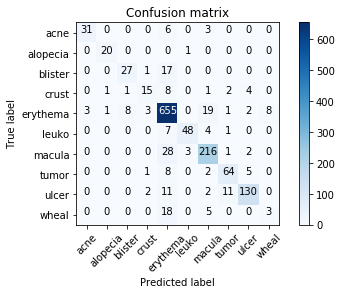}
	\caption{Confusion matrix for the quality of evaluation on dermatological dataset classified
		by learning on a pre-trained ResNet-152}
	\centering
\end{figure}

\begin{table*}[!t]
	\caption{Performance comparison between training schemes on dermatological images}
	\vspace{-0.8em}
	\begin{center}
		\begin{tabular}{c | c c | c c| c}
			\hline
			\hline
			\makebox{Model} & {\small Accuracy (Conventional)} & {\small Time (s)} & {\small Accuracy (DLR+CLM)} 
			& {\small Time (s)} & {\small Speedup}\\
			\hline
			ResNet-34		&	72.62\%	&   1576.33	&   83.40\%	& 343.20    & 4.6 \\
			ResNet-50		&	71.88\%	&	2130.27	&   83.26\%	& 692.64 	& 3.1 \\
			ResNet-101		&	71.70\%	&	4216.22	&   83.19\%	& 1076.90 	& 3.9 \\
			ResNet-152		&   71.56\%	&	5437.72 &   84.90\%	& 958.88 	& 5.7 \\
			DenseNet-161    &   78.02\%	&   4156.75	&   84.61\%	& 946.14 	& 4.4 \\
			\hline
			\hline
		\end{tabular}
		\label{derma-optim}
	\end{center}
\end{table*}

\section{Conclusion}
We have demonstrated that there is much room for performance improvement in classical techniques 
for small datasets, by tuning the learning rate. It not only speeds up the convergence, 
but allows better model fits, increasing validation accuracy. By slowing down learning in initial 
layers of CNN, it is easy to shift the bulk of computation towards the final layers, resulting in 
optimization of the process. Further, we observed that because of this combination of established 
techniques, models are more immune to learning bias in an unbalanced dataset, than traditional methods. 

\section*{Acknowledgement}
The authors offer their sincere thanks to D. Page (Myrtle AI) and FastAI community, 
who offered various tools \& resources for easy interpretation.

\newpage
\onecolumn
\section*{Appendix}

\begin{figure}[ht] 
  \begin{subfigure}[b]{0.5\linewidth}
    \centering
    \includegraphics[width=0.95\linewidth]{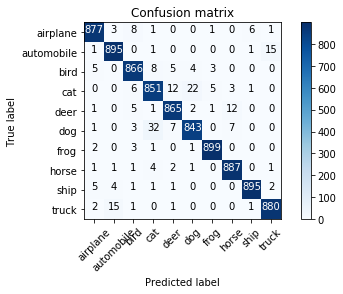} 
    \caption{ResNet-34} 
    \vspace{4ex}
  \end{subfigure}
  \begin{subfigure}[b]{0.5\linewidth}
    \centering
    \includegraphics[width=0.95\linewidth]{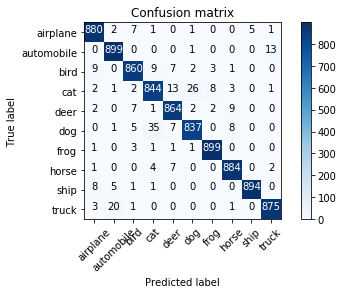} 
    \caption{ResNet-50} 
    \vspace{4ex}
  \end{subfigure} 
  \begin{subfigure}[b]{0.5\linewidth}
    \centering
    \includegraphics[width=0.95\linewidth]{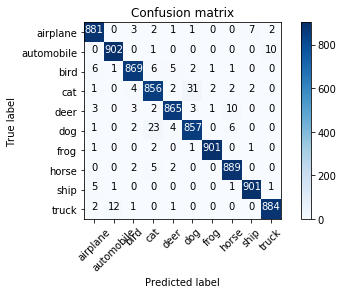} 
    \caption{ResNet-101} 
  \end{subfigure}
  \begin{subfigure}[b]{0.5\linewidth}
    \centering
    \includegraphics[width=0.95\linewidth]{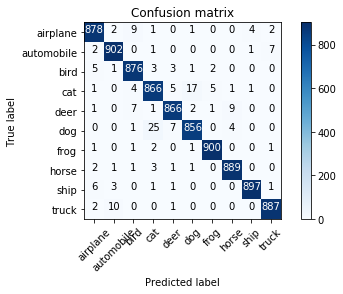} 
    \caption{ResNet-152} 
  \end{subfigure}
  \caption*{Confusion matrices of validation set from different ResNet 
  architectures trained on CIFAR-10 dataset}
\end{figure}

\end{document}